\documentclass{article}

\usepackage[utf8]{inputenc} 
\usepackage[T1]{fontenc}    
\usepackage{hyperref}       
\usepackage{url}            
\usepackage{booktabs}       
\usepackage{amsfonts}       
\usepackage{nicefrac}       
\usepackage{microtype}      
\usepackage{xcolor}         
\usepackage[most]{tcolorbox}
\usepackage{mathtools}
\usepackage{thmtools}
\usepackage[normalem]{ulem}

\usepackage{amsmath,amssymb,amsthm}
\usepackage{tikz}
\usetikzlibrary{positioning}
\usepackage{graphicx}
\usepackage{framed}
\usepackage[svgnames]{xcolor}
\usepackage{geometry}
\usepackage{stmaryrd}
\usetikzlibrary{positioning, arrows.meta, shadows.blur}

\definecolor{shadecolor}{named}{LightGray}
\newcommand{\todo}[1]{\textbf{\textcolor{red}{TODO: #1}}}

\newcommand{\tcs}{{\textsc{TCS-Bench}}}

\usepackage{comment}

\title{TCS-BENCH: Benchmarking State-of-the-Art Generative AI  Theoretical Computer Science Research Ability}

\author{
  Vincent Cohen-Addad\footnote{co-First author. Names appear in alphabetical order}\;\textsuperscript{,}\footnote{Google 1}  \and
  Dimitris Paparas \footnotemark[1]\;\textsuperscript{,}\footnotemark[2] \and Ernest van Wijland\footnotemark[1]\;\textsuperscript{,}\footnote{CNRS, IRIF, Université Paris-Cité}\;\textsuperscript{,}\footnotemark[2] \and
  Max Springer\footnote{Princeton University, Department of Computer Science} \and
  Julien Canitrot-Paradis\footnote{Université Paris-Saclay, CEA, List, Palaiseau, France} 
  \and Honghao Lin\footnotemark[2]
  \and David P. Woodruff\footnotemark[2]
  \and
  Adarsh Kumarappan\footnotemark[2]\;\textsuperscript{,}\footnote{California Institute of Technology, Department of Computer Science} \and
  Rajesh Jayaram\footnotemark[2] \and
  Rudrajit Das\footnotemark[2] \and
  Lalit Jain\footnotemark[2] \and
  Ola Svensson\footnotemark[2] \and
  Silvio Lattanzi\footnotemark[2] \and
  Mislav Balunovic\footnotemark[2] \and
  Theophane Weber\footnotemark[2] \and
  Vahab Mirrokni\footnotemark[2] 
}
\newcommand{\colosseum}{{\textsc{Colosseum}}}
\date{}

\begin{document}

\maketitle

\begin{abstract}
  We introduce \tcs, a benchmark for evaluating Large Language Models (LLMs) on research-level Theoretical Computer Science  (TCS) proof generation.
\tcs \ consists of theorem-proving tasks from papers published at top theoretical computer science venues (STOC, FOCS, and SODA). Each task provides the necessary context to derive a self-contained proof for a target result. We evaluate state-of-the-art models on this benchmark.
We verify the correctness of generated proofs via a verification agent, and further benchmark the verifier against human-expert proof judgements on a set of target statements and generated proofs pairs. 
Our reference verifier achieves over 90\% accuracy on the expert labeled set.
The tasks are publicly available at \url{https://github.com/TCSBench/TCSBench}.
\end{abstract}

\section{Introduction}
Over the last two years, Large Language Models (LLMs) have achieved superhuman performance on a variety of standardized benchmarks, from professional exams to programming competitions~\cite{achiam2023gpt,alphaproof2024ai, dekoninck2025open, kung2023performance, trinh2024solving,varanasi2023gpt,zhong2023agieval}. 
These successes have extended into the formal domain of mathematics, where specially primed models have demonstrated the ability to solve problems at the level of an International Mathematical Olympiad (IMO) gold medalist~\cite{chervonyi2025gold}, suggesting that LLMs can emulate rigorous mathematical logic. 

Recently, these capabilities have crossed the threshold into open-ended scientific discovery with models actively contributing to expert-level mathematical and scientific breakthroughs. For example, Google's Gemini Deep Think and its advanced variants have successfully collaborated with human researchers to solve open problems, refute conjectures, and generate novel proofs across theoretical computer science, optimization, and physics \cite{woodruff2026acceleratingscientificresearchgemini},\cite{feng2026autonomousmathematicsresearch}.
An internal version of OpenAI's ChatGPT disproved the unit-distance conjecture \cite{openai2026unit} and other long-standing mathematical and theoretical computer science problems \cite{openai2026tenadv}. Anthropic's models have similarly been utilized to independently discover counterexamples to long-standing conjectures \cite{anthropic2026}.

However, these frontier-level successes highlight a critical limitation in how the AI community evaluates language models: a significant gap has emerged between what state-of-the-art agents can achieve in active research and what our standardized benchmarks can actually measure. 
We find the research field to be a rich, sustainable, and challenging testbed for evaluating the next generation of language models.
Existing benchmarks, while valuable, fail to capture the core difficulties of real-world mathematical research for several key reasons.
First, competition problems like those found in the IMO are typically self-contained,
whereas research theorems are deeply embedded in the context of bespoke definitions, notations, and previously established lemmas.
Second, research results frequently involve constructing a scaffold of interconnected results, a task that goes far beyond finding a single, clever insight.
To truly measure progress, we need challenging benchmarks that more accurately reflect the work of human researchers.

\subsection{Our Contributions}
To this end, we introduce \tcs{}, a new benchmark for evaluating an LLM's ability to prove theorems from cutting-edge Theoretical Computer Science (TCS) research papers.
The task is grounded in scientific practice: a model is presented with a target statement extracted from a paper published at a top-tier conference, and is tasked with generating a proof. The challenging part is to provide all required ``basic'' mathematical context required to prove the target statement. Our key contribution is to provide a self-contained task that can be solved without access to the internet, and that evaluates the ability of the models to come up with a proof from first principles and the intermediate, state-of-the-art lemmas provided in the context. This is obtained by processing
the paper the task is extracted from and the target proof. 

Furthermore, a central aspect in our task generation process is the ability to generate harder and harder tasks by having a tight control over the generated context. Indeed, by masking intermediate Lemmas used in the proof of the target statement, the task becomes harder.
One can thus increase the difficulty of the tasks by simply masking more and more intermediate results, all the way to define tasks that consist in proving the main result of the paper.

Then successfully resolving a task requires a model to comprehend the surrounding context, understand the intricate connections between lemmas and theorems, and generate a logically sound  proof that would
provide a similar amount of details as the peer-reviewed proof extracted from the paper.
This format challenges models to perform the context-dependent reasoning that is a hallmark of scientific discovery.
Our contributions are as follows:
\begin{itemize}
    \item \tcs, a benchmark composed of 300 theorem-proving tasks from papers published at the top theoretical computer science venues, FOCS, STOC, and SODA, between 2020 and 2026. We make the tasks publicly available at \url{https://github.com/TCSBench/TCSBench}.
    \item We develop and validate an automated proof verification system, achieving 
    more than $90\%$ accuracy against human expert judgments on a held-out set of 100 human-labeled proofs.
\end{itemize}

\paragraph{Task Construction Overview.}

The foundation of our benchmark
is a curriculum of statement-proving tasks from publicly available conference proceedings and that can and should be solved without requiring access to the internet.
At a high level, we scrape and process the LaTeX source of a paper, build the dependency graph of the statements (theorems, lemmas, claims, etc), and for each of the statements we assemble multiple versions of the context needed to attempt the proof, hiding varying subsets of dependencies.
Each task is then packaged as a self-contained JSON entry comprising three fields: a context, a target statement and a ground-truth proof.
Full technical details of the benchmark construction are given in Section~\ref{sec:benchmark-construction}.

\paragraph{Benchmark Overview.}
Tasks are selected to span a broad range of TCS subfields and, critically, a broad range of difficulty levels.
This stratification ensures that \tcs\ is informative across the full capability spectrum of current and future models, rather than saturating at either extreme.
A model is evaluated on \tcs\ by generating a proof for each task in the benchmark and submitting those proofs to an automated (and validated) verifier.
We specify a reference verifier and report all our baseline results using this to ensure reproducibility.


\section{Related Work}
Our work is situated at the intersection of benchmarks for mathematical reasoning, formal proof generation, and the broader evaluation of AI in scientific research workflows.

\paragraph{Competition Mathematics Benchmarks.}
A significant body of work has focused on evaluating LLMs on competition-style mathematics problems.
Benchmarks in this area often draw from sources like the International Mathematical Olympiad (IMO), testing a model's ability to find clever insights for well-defined, self-contained problems~\cite{balunovic2025matharena,dekoninck2025open,liu2025combibench,mao2024champ,trinh2024solving}.
A notable success in this domain is AlphaGeometry2, which demonstrated performance equivalent to an IMO gold medalist~\cite{huang2025gemini}. 
While these benchmarks are invaluable for measuring discrete problem-solving abilities, their self-contained nature does not reflect the challenges of genuine research~\cite{raji2021ai}.
In contrast, \tcs{} sources its problems directly from published literature.
This requires the model not just to solve a problem, but to reason within the rich, interdependent theoretical context established by the paper's definitions, notations and prior lemmas.

\paragraph{Research-Level Mathematical Benchmarks.}
Recent work has begun to address the gap between competition mathematics and genuine research.
HorizonMath~\cite{wang2026horizonmath} presents over 100 predominantly unsolved problems from computational and applied mathematics, leveraging a generator-verifier gap where solutions are hard to produce but efficient to verify through numerical comparison or deterministic constraint checking.
LemmaBench~\cite{peyronnet2026lemmabench} takes a complementary approach by automatically extracting and contextualizing lemmas from recent arXiv preprints, creating an updatable benchmark immune to contamination through continuous refreshment from new publications.
BrokenMath~\cite{petrov2025brokenmath} evaluates a different failure mode—sycophancy in theorem proving—by perturbing valid mathematical statements into plausible but false versions, revealing that even frontier models attempt to prove false statements 29-70\% of the time.
\tcs{} shares with these works the focus on research-level mathematics beyond competition problems, but differs in targeting proof completion within the rich contextual dependencies of published theoretical computer science papers, where success requires not just solving isolated problems but reasoning within an interconnected scaffold of definitions and prior results.

\paragraph{Formal Theorem Proving.}
Another major direction of research focuses on benchmarking LLMs for proof generation in formal languages using interactive theorem provers like Lean or Coq~\cite{alphaproof2024ai,bayazit2025case,dekoninck2025open,lama2024benchmarking,li2025proving}.
This work evaluates a model's ability to generate sequences of tactics to construct a machine-verifiable proof, often within established formal mathematics libraries.
Similarly, benchmarks like MathConstruct test a model's ability to generate a specific mathematical object whose properties can be formally verified by an automated function~\cite{dekoninck2025mathconstruct}.
This line of work is critical for ensuring logical rigor. \tcs{} complements these efforts by focusing on a different but equally important task: generating complete, human-readable proofs in natural language.
The evaluation challenge in our work shifts from formal machine-verifiability to assessing logical soundness within the implicit argumentative structure of a research paper—a task that is more aligned with the daily workflow of human mathematicians.

\paragraph{AI for Scientific Discovery.}
As LLMs advance, the focus of evaluation is shifting from solving established problems to assessing their potential to accelerate scientific discovery~\cite{feng2026autonomousmathematicsresearch,gottweis2025towards,jain2023gflownets,wang2023scientific,woodruff2026acceleratingscientificresearchgemini,zhang2025exploring}. 
Recent work has demonstrated their utility across numerous domains, from chemistry and biology to astrophysics~\cite{boiko2023autonomous,huang2024crispr,irwin2022chemformer,parker2024astroclip,vinuesa2023transformative}.
Our work aligns with a new class of benchmarks designed to evaluate AI agents on complex, long-horizon scientific tasks.
For example, PaperBench evaluates an agent's ability to replicate an entire AI research paper, a process that includes understanding the paper, developing a codebase, and executing experiments to reproduce its empirical results~\cite{staracepaperbench}.
This paradigm assesses a model's practical utility in a realistic research workflow.
\tcs{} adapts this ``research-as-benchmark'' paradigm to the domain of theoretical computer science.
While PaperBench focuses on replicating empirical results, \tcs{} is the first to focus on replicating theoretical contributions.
By isolating the task to proof synthesis within the context of a research paper, our benchmark provides a targeted measure of the sophisticated, context-dependent reasoning required to contribute to the frontiers of mathematical knowledge.

\section{The TCS Benchmark}
\label{sec:benchmark-construction}

\tcs{} evaluates a model's ability to prove theorems drawn from cutting-edge theoretical computer science research.
Each task in the benchmark is a self-contained proof-completion problem wherein the model receives a curated context comprised of definitions, prior lemmas and condensed external references.
Given this primer knowledge together with a target statement, the model is tasked with producing a complete proof.
Crucially, all tasks are derived from published papers whose proofs have been systematically removed, so that success requires chains of complex mathematical reasoning rather than simple memorization and recall.

\subsection{Data Acquisition and Preprocessing}
We construct the benchmark from papers published at FOCS, STOC, and SODA, between 2020 and 2026.
For each paper, we obtain the \LaTeX{} source from arXiv, limiting our dataset to papers released under permissive licenses (CC-0 or CC-BY-4.0)  to ensure all benchmark content is legally distributable. We list all papers we used in appendix \ref{sec:papers}.

Research proofs routinely invoke results from prior literature. To ensure tasks are self-contained, we skip target statements whose proofs invoke external results without restating them.


\subsection{Dependency Structure Construction}
The core of our benchmark construction is a structural analysis that extracts the logical dependency structure of each paper. 
This process produces a directed acyclic graph (DAG) over all formal statements, where an edge from statement A to statement B indicates that the proof of B depends on A.

To extract this graph, we first use a deterministic LaTeX parser to identify all theorem-like environments (theorem, lemma, definition, corollary, etc.) and assign each a unique identifier. We then employ an LLM-based analysis pass to:
\begin{itemize}
    \item Map each proof environment to the statement it proves (non-trivial since proofs may appear out of order or span multiple environments)
    \item Analyze the full paper to construct dependency edges—identifying when the proof of statement B invokes statement A
\end{itemize}

We programmatically verify acyclicity and reject any paper for which the extracted graph contains a cycle, as this indicates an extraction error.

From the DAG, we compute the \textbf{rank} of each statement as the length of the longest directed path terminating at that node. Definitions and axioms have rank 0, and each subsequent layer of derived results increments the rank. This ranking serves two purposes: it stratifies tasks by difficulty (higher-rank proofs require reasoning about longer chains of dependencies), and it prevents information leakage during context assembly (we can systematically hide all results of rank $\ge r$ when constructing a task for a rank-$r$ statement).

\begin{figure}[t]
  \centering
  \resizebox{\textwidth}{!}{%
  \begin{tikzpicture}[
      >=Stealth,
      node distance=8mm and 12mm,
      dagnode/.style={
        draw=gray!40,          
        fill=gray!5,           
        line width=0.6pt, 
        align=left, 
        inner sep=8pt, 
        minimum width=44mm, 
        text width=42mm, 
        font=\small,
        rounded corners=4pt
      },
      toprove/.style={
        draw=blue!70,          
        fill=blue!4,           
        line width=1.5pt       
      },
      edge/.style={
        ->,                    
        draw=gray!60,
        thick
      }
    ]

    \node[dagnode, toprove] (thm) {%
      \textbf{Theorem T} \hfill {\scriptsize\color{gray} rank 3}\\[1mm]
      {\scriptsize \emph{Statement:} ``\dots''}\\
      {\scriptsize \emph{Deps:} \{Lemma C\}}%
    };

    \node[dagnode, toprove, below=of thm] (l3) {%
      \textbf{Lemma C} \hfill {\scriptsize\color{gray} rank 2}\\[1mm]
      {\scriptsize \emph{Statement:} ``\dots''}\\
      {\scriptsize \emph{Deps:} \{Lemma A, Lemma B\}}%
    };

    \node[dagnode, below left=8mm and 6mm of l3] (l1) {%
      \textbf{Lemma A} \hfill {\scriptsize\color{gray} rank 1}\\[1mm]
      {\scriptsize \emph{Statement:} ``\dots''}\\
      {\scriptsize \emph{Deps:} \{Def.~1\}}%
    };
    
    \node[dagnode, toprove, below right=8mm and 6mm of l3] (l2) {%
      \textbf{Lemma B} \hfill {\scriptsize\color{gray} rank 1}\\[1mm]
      {\scriptsize \emph{Statement:} ``\dots''}\\
      {\scriptsize \emph{Deps:} \{Def.~1, Def.~2\}}%
    };

    \node[dagnode, below=of l1] (d1) {%
      \textbf{Def.~1} \hfill {\scriptsize\color{gray} rank 0}\\[1mm]
      {\scriptsize \emph{Statement:} ``\dots''}\\
      {\scriptsize \emph{Deps:} $\varnothing$}%
    };
    
    \node[dagnode, below=of l2] (d2) {%
      \textbf{Def.~2} \hfill {\scriptsize\color{gray} rank 0}\\[1mm]
      {\scriptsize \emph{Statement:} ``\dots''}\\
      {\scriptsize \emph{Deps:} $\varnothing$}%
    };

    \draw[edge] (thm) -- (l3);
    \draw[edge] (l3) -- (l1);
    \draw[edge] (l3) -- (l2);
    \draw[edge] (l1) -- (d1);
    \draw[edge] (l2) -- (d1);
    \draw[edge] (l2) -- (d2);
    
  \end{tikzpicture}
  }
  \caption{DAG example}
  \label{fig:DAGexample}
\end{figure}
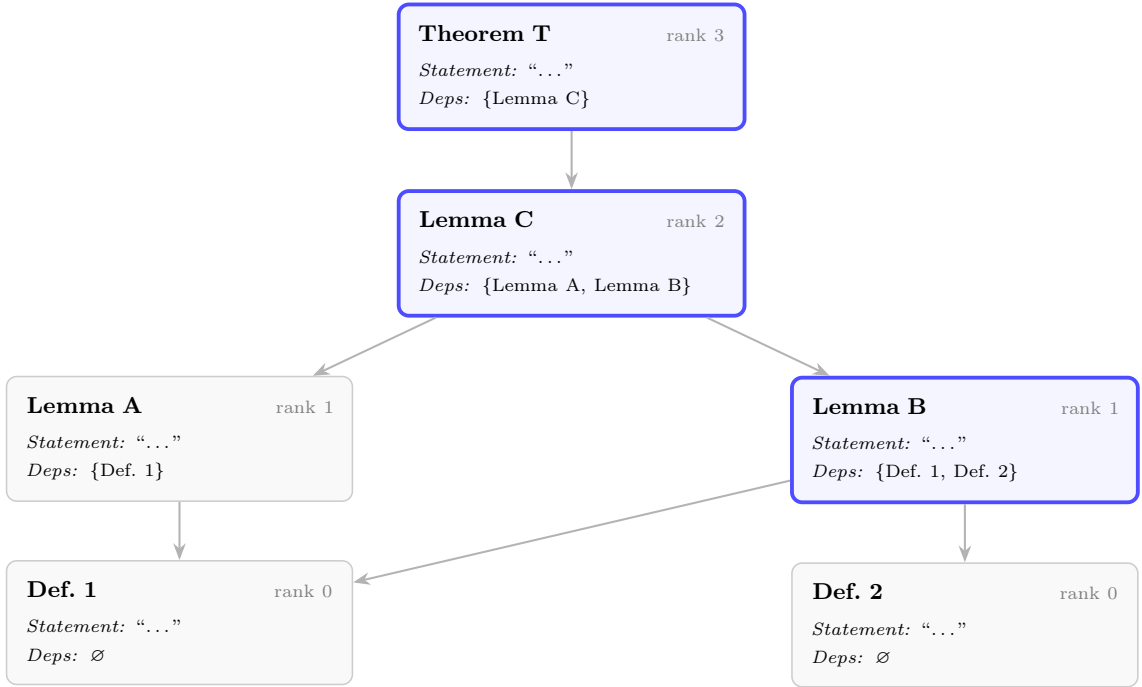

Figure~\ref{fig:DAGexample} illustrates an example dependency DAG extracted from a paper. 
The graph shows how a main theorem (rank 3) depends on intermediate lemmas (ranks 1-2), which in turn depend on foundational definitions (rank 0). 
Each blue node represents a proof-completion task in our benchmark. Note that only statements that have corresponding \texttt{$\backslash$begin\{proof\}$\backslash$end\{proof\}} tags in the paper are turned into tasks for our benchmark.

\subsection{Task Construction}
Each proof-completion task consists of three components: a \emph{context}, a \emph{target statement}, and a \emph{ground-truth proof} (withheld during evaluation).
The key challenge in constructing high-quality tasks is producing a context that is both self-contained (containing all information logically necessary to derive the proof) and concise enough to fit within standard context windows.
We achieve this through the following procedure.

\paragraph{Initial Context Assembly.}
For a target statement $s$ of rank $r$, we initialize the context by concatenating $(i)$ all resolved external reference digests, $(ii)$ the paper's text truncated at the start of $s$'s proof, with two categories of redaction applied: all statements of rank $\ge r$ are hidden to prevent information leakage from later results, and all proof environments are removed.
If any dependency of $s$ falls outside the truncated portion of the paper, we re-insert its statement into the context.

\paragraph{Scalable Difficulty Tasks.}

To create tasks spanning a range of difficulties, we exploit the dependency structure to generate multiple variants of each proof task.
Starting from a base task where all dependencies are provided in the context, we systematically withhold intermediate results, requiring the model to discover and prove them on the way to the main target.

\begin{figure}[t]
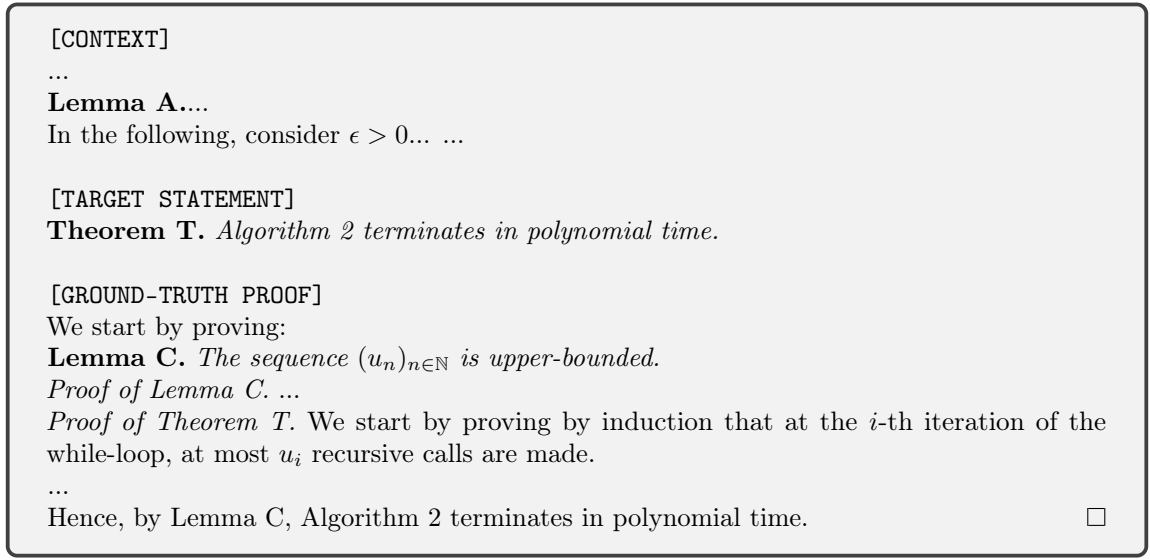

\centering
\begin{tcolorbox}
\noindent\texttt{[CONTEXT]}

...

\noindent\textbf{Lemma A.}...

\noindent In the following, consider $\epsilon > 0$...
...
\\

\noindent\texttt{[TARGET STATEMENT]}

\noindent\textbf{Theorem T.} \textit{Algorithm 2 terminates in polynomial time.}\\

\noindent\texttt{[GROUND-TRUTH PROOF]}

We start by proving:

\noindent\textbf{Lemma C.} \textit{The sequence $(u_n)_{n\in\mathbb{N}}$ is upper-bounded.}

\noindent\textit{Proof of Lemma C.} ...

\noindent\textit{Proof of Theorem T.} We start by proving by induction that at the $i$-th iteration of the while-loop, at most $u_i$ recursive calls are made.

...

Hence, by Lemma C, Algorithm 2 terminates in polynomial time.
\qed
\end{tcolorbox}
\caption{Task with omitted intermediary results to increase complexity.}
\label{fig:tcs-bench-hard}
\end{figure}

\vspace{2.5mm}
\noindent \emph{Example.}
Consider proving Theorem T from Figure~\ref{fig:DAGexample}, which depends on Lemma C, which in turn depends on Lemmas A and B.
The base difficulty for prompting a model to construct a proof would be to supply the model with all dependent results within the context (as depicted in Figure~\ref{fig:tcs-bench-easy}). 
Furthermore, we can easily increase the task complexity by omitting a subset of the dependencies (example prompting in Figure~\ref{fig:tcs-bench-hard}).
Thus, forcing the model to prove such intermediary results along the way to the final claim.
This procedure generates a spectrum of difficulties: at one extreme, all dependencies are provided and the model need only combine them; at the other, the model must reconstruct substantial portions of the paper's proof architecture. 

\begin{figure}[t]
\begin{tcolorbox}
\noindent\texttt{[CONTEXT]}

...

\noindent\textbf{Lemma A.}...

\noindent\textbf{Lemma C.} \textit{The sequence $(u_n)_{n\in\mathbb{N}}$ is upper-bounded.}

\noindent In the following, consider $\epsilon > 0$...
...
\\

\noindent\texttt{[TARGET STATEMENT]}

\noindent\textbf{Theorem T.} \textit{Algorithm 2 terminates in polynomial time.}\\

\noindent\texttt{[GROUND-TRUTH PROOF]}

\noindent\textit{Proof.} We start by proving by induction that at the $i$-th iteration of the while-loop, at most $u_i$ recursive calls are made.

...

Hence, by Lemma C, Algorithm 2 terminates in polynomial time.
\qed
\end{tcolorbox}
\caption{Base difficulty task (all dependencies provided)}
\label{fig:tcs-bench-easy}
\end{figure}


\paragraph{Context Compression.}
To make the tasks short enough to fit the standard context window of $10,000$ tokens, we apply the following procedure.

First, we conduct \emph{iterative section pruning}.
Since many papers contains sections (e.g. related work, motivating context) that are irrelevant to a given target statement,
we iteratively parse the section hierarchy of the assembled context and prompt an LLM to identify sections/subsections/subsubsections that are entirely irrelevant to the target problem and its proof.
Identified sections are removed, and the process repeats until no further pruning is possible.
Second, we apply an \emph{LLM shortner} which, following section pruning condenses the remaining context (ie. shortening verbose passages or tightening exposition) while preserving all mathematically essential content.


\subsection{Quality Filtering}

To address potential information loss or artifacts introduced during context assembly, we implement a final quality filtering stage to ensure all generated tasks are self-contained, well-posed, and free of information leakage. This filtering applies two categories of automated criteria. First, structural checks exclude tasks that contain text-incompatible elements (such as figure references), unresolvable dependencies, leaked proof metadata, or excessive token lengths. Second, semantic checks utilize large language models to verify that all mathematical objects are properly defined, the target statement is unambiguous and correctly proven by the ground truth, and the assembled context remains logically and mathematically coherent. Any task failing these structural or semantic checks is excluded from the final benchmark.

\subsection{Task Formulation Summary}
Each released task provides the solver with:
\begin{enumerate}
    \item A self contained \LaTeX{} context ($\le$ 10,000 tokens) comprising definitions, prior results (with proofs omitted), and condensed external reference digests.
    \item A target statement to be proved.
\end{enumerate}

The solver must produce a single, complete proof of the target statement, using \LaTeX\ for mathematical notations.
All justifications must follow from results present in the provided context and the solver is prohibited from accessing the original paper or external sources.
The ground-truth proof is withheld and used exclusively for the evaluation.

\section{Automated Proof Verification} \label{sec:verifier}
A primary challenge in a benchmark like \tcs{} is the need for a scalable and reliable method to evaluate the correctness of generated mathematical proofs.
Manual verification by human experts is prohibitively slow and expensive.
To address this, we developed a specialized verifier agent, tasked with deciding whether a candidate solutions constitutes a valid proof.

\subsection{Verifier Design}
The verifier's goal is to make a binary decision (correct or incorrect) on a candidate proof for a specific statement within a given context.
It is aligned to tolerate trivial omissions common in academic literature (e.g., ``the rest follows by simple algebra'') but reject proofs with critical logical gaps or errors.

The verifier is provided with three key inputs: the task's context and target statement that were passed to the solver, and additionally the ground-truth proof.
Then, four calls are made to Gemini 3.1 Flash, a cheap model, and a candidate proof is deemed correct if and only if at least three of the four verdicts mark it as correct. 
{The prompt instructs the model to rigorously assess the submission across four dimensions: logical rigor (ensuring strict adherence to provided lemmas without introducing unauthorized assumptions), precise variable tracking and scoping, semantic integrity (penalizing hand-waving or jargon-heavy hallucinations), and exact quantitative accuracy in derivations and boundary conditions.}

\subsection{Verifier Calibration}

To design the verifier prompt, we generated proofs by running the solver on a set of tasks, disjoint from the benchmark. Then, human experts reviewed them, and produced a set of 50 correct proofs and 50 incorrect proofs.
Finally, we ran the GEPA~\cite{agrawal2025gepa} improvement pipeline on this alignment task to produce a prompt that achieves an accuracy of more than $90\%$.

\section{Experimental Setup}

We evaluate frontier language models on \tcs\ to establish baseline performance on research-level mathematical proof generation. 
Our evaluation focuses on measuring the current capabilities of state-of-the-art models when presented with proof-completion tasks drawn from cutting-edge theoretical computer science research.

\subsection{Experimental Setup}
We assess the following frontier models: Gemini 3.1 Pro and Gemini 3.1 DeepThink, Opus 5, and GPT 5.6 Pro. We also present \colosseum, an internal harness that we evaluate using Gemini 3.1 Pro, Gemini 3.7 Flash, and a combination of the two.
All models were accessed via their respective API endpoints or web interfaces using the most recent versions available at the time of evaluation.
For each task, we query the model with the context and target statement as described in Section~\ref{sec:benchmark-construction}. We provide the exact prompt formatting in appendix \ref{sec:prompts}.

For models offering extended reasoning capabilities, we use the maximum publicly available thinking budget to allow models to fully explore the problem space.
%
Generated proofs are evaluated using our automated verifier (Section~\ref{sec:verifier}). 
Each model attempts all 300 tasks in \tcs.

\paragraph{Results \& Analysis.}
Table~\ref{table:results} presents the overall performance of each model on \tcs.
The strongest performing model, GPT 5.6 Pro, achieves an accuracy score of $68\%$, successfully proving 204 of 300 tasks. Notably, Opus 5, exhausts its token budget of 128K tokens before obtaining an answer for 162 out of the 300 tasks, which negatively affects its performance.

\begin{table}[t]
\centering
\caption{Model performance on TCS-BENCH measured by accuracy.}
\label{table:results}
\begin{tabular}{lc}
\toprule
\textbf{Model} & \textbf{Accuracy} ($\shortuparrow$) \\
\midrule
Opus 5 & 32.77 \\
Gemini 3.1 Pro & 30.3 \\
Gemini 3.1 DeepThink & 52 \\
GPT 5.6 Pro (max) & 68 \\
Colosseum Cross Model & 71 \\
\bottomrule
\end{tabular}
\end{table}

\paragraph{Results \& Analysis with \colosseum}
\label{sec:colosseum}

The results above evaluate base models directly. We also report results for
\colosseum, an agentic proof-search harness we run on top of a base model: for
each problem it explores several candidate proof strategies, decomposes the
target statement into subproblems, solves them, and assembles and revises a
final proof. We do not describe \colosseum{} in detail here; the only property
that matters below is that it is parameterised by its base model, so the same
pipeline can be run on different models to obtain independent proofs of the
same problem.

\paragraph{Selecting between two runs.}
Running \colosseum{} on two different base models yields two independent proofs
of every problem, and we select one of them automatically. The selection asks
the second base model to \emph{critique} the first model's proof: we draw $8$
independent critiques and submit the second model's proof if at most half of
them judge the first model's proof to be correct, and the first model's proof
otherwise.

This critic is a component of \colosseum \ and is distinct from the automated
grader of Section~\ref{sec:verifier}. The grader is used only to score final
submissions; the selection has no access to it, to ground truth, or to any
human judgement. 

\paragraph{Why cross-model.}
The rule relies on one observation: a model is a poor judge of its own proof,
but a competent judge of another model's. On \tcs{}, $93.4\%$ of the cases in
which Gemini 3.1 Pro accepts its own proof are unanimous, whereas cross-model
critiques separate correct from incorrect proofs well (AUC $0.854$--$0.896$).
The result is not sensitive to the precise critique cutoff. Across cutoffs
from $0/8$ through $7/8$, accuracy remains between $68.3\%$ and $71.0\%$.

\begin{table}[t]
\centering
\begin{tabular}{lcc}
\toprule
& Solved / 300 & Accuracy \\
\midrule
\colosseum, Gemini 3.1 Pro                       & 162 & $54.0\%$ \\
\colosseum, Gemini 3.7 Flash               & 165 & $55.0\%$ \\
\colosseum, cross-model selection between the two & 213 & $71.0\%$ \\
\midrule
Oracle best-of-two (upper bound, not achievable) & 217 & $72.3\%$ \\
\bottomrule
\end{tabular}
\caption{\colosseum \ run on two different base models, with automated
cross-model selection between the resulting proofs. All rows are graded by the
automated grader of Section~\ref{sec:verifier}. The oracle row reports the
fraction of problems solved by at least one of the two runs; it requires
knowing the answer and upper-bounds any selection rule.}
\label{tab:colosseum}
\end{table}





While we lack comprehensive human baselines, we note that all tasks in TCS-BENCH have ground-truth proofs published by domain experts. 
The gap between the strongest model $71\%$ and perfect performance (100\%) represents the current frontier of automated mathematical reasoning.




\section{Discussion}
We have introduced \tcs{}, the first benchmark evaluating LLMs on proof generation from cutting-edge theoretical computer science research. 
Unlike competition mathematics benchmarks testing isolated problems, \tcs{} requires reasoning within the rich contextual scaffolding of research papers—navigating bespoke definitions, dependency structures, and chains of intermediate results across 300 tasks spanning multiple difficulty levels.

A key contribution is our automated proof verification system achieving over 90\% accuracy against human expert judgments, enabling scalable evaluation without prohibitive manual verification costs. 
The verifier tolerates stylistic variations common in mathematical writing while rigorously detecting logical gaps—a balance essential for meaningful evaluation.

Our evaluation of five frontier models shows the strongest systems: Gemini 3.1 DeepThink  solves 52\%, GPT-5.6-Pro solves 68\% and Colosseum with a cross-model approach solves 71.0\%. 
While this demonstrates meaningful progress in automated mathematical reasoning, the gap to perfect performance reveals substantial room for advancement. 
The stable performance ceiling across all models suggests current architectures face fundamental limitations in constructing multi-step mathematical arguments within complex dependencies. 

We lastly note that \tcs{} is designed for longevity through continual addition of new papers that postdate model training cutoffs, scalable difficulty via dependency-hiding that spans easy to extremely challenging variants, and rank-based stratification enabling fine-grained progress tracking. 

\bibliographystyle{plain}
\bibliography{references}


\clearpage
\appendix
\section{Omitted Details}
\subsection{Full Prompt Details}
\label{sec:prompts}
We present here the prompt formatting for requesting the evaluated model to prove a target statement:

\begin{shaded}
\begin{ttfamily}
You are an expert mathematician. Your task is to provide a thorough and correct proof for a mathematical problem (your Target Problem). This proof will be used to benchmark your abilities as a mathematician.
\\
\\
To obtain your proof, you can assume and use any reference (lemma, theorem, definition, etc) from the context below:
\\
\\
********** BEGIN CONTEXT **********
\\
\\
\{CONTEXT\}
\\
\\
********** END CONTEXT **********
\\
\\
********** BEGIN EVALUATION CRITERIA **********
\\
\\
Recall that you can assume and use any of the mathematical statements provided in the context above.
\\
\\
Your proof must be complete and valid. As a guideline, here is a non-exhaustive list of criteria to evaluate your proof. You should use it, together with any additional criteria you can think of, to evaluate any candidate proof before your final response.
\\
\\
Evaluation Criteria:
\\
\\
1. **Logical Rigor and External Assumption Check**:
\begin{itemize} \item[-] **No Unauthorized Constraints**: The proof must not introduce arbitrary numerical constraints or "safety margins" to simplify the proof (e.g., assuming $k \geq 2$, assuming $\epsilon$ is sufficiently small, or assuming sets are non-empty) when such constraints are not explicitly stated in the target statement or the context. 
\item[-] **Strict Lemma Adherence**: Every step must follow strictly from the provided context or standard mathematical knowledge. Applying a theorem without explicitly verifying all its preconditions (as defined in the context) is a failure.
\item[-] **Case Integrity**: If the proof involves case-splitting (e.g., $i^* \leq T$ vs $i^* = T+1$), the proof must address these specific boundaries. Skipping a boundary case or "merging" distinct logical paths via hand-waving results in a failure.
\end{itemize}
2. **Variable Alignment and Property Scope**:
\begin{itemize}
\item[-] **Property Scope Integrity**: Properties must only be applied to the specific variables for which they are defined. If the proof generalizes a property of a subset to a larger set then this constitutes a failure (e.g., applying a $u$-uniformity property defined for "outer queries" $T_2$ to the "total queries" $T$, or applying a property of a specific index $j$ to all indices $i$ without proof).
\item[-] **Index and Set Precision**: The proof must maintain the integrity of sets (e.g., $B$ vs $B^*$, $S_i$ vs $S_{{i-1}}$) and indices. Misidentifying a variable or applying a property to the wrong time step or set is a fatal error.
\end{itemize}
3. **Semantic Integrity and Jargon Detection**:
\begin{itemize}
\item[-] **No Hallucinated Logic/Word Salad**: If the proof uses repetitive, nonsensical, or overly dense jargon to mask a lack of logical depth, then this is a failure. The proof must be linguistically coherent. If a paragraph consists of technical terms strung together without clear propositional logic (e.g., "mapping limits matching identical parameters constraints mapping"), it is a failure.
\item[-] **No "Standard" Hand-waving**: The proof cannot skip non-trivial derivations by claiming they are "standard", "trivial", etc.
\end{itemize}
4. **Quantitative and Limit Accuracy**:
\begin{itemize}
\item[-] **Derivation Accuracy**: Any error in arithmetic, algebraic manipulation, or inequality direction will automatically invalidate the proof.
\item[-] **Boundary and Floor/Ceiling Precision**: Bounds must be exactly supported. For example, if a floor function $\lfloor k^{{-\epsilon}} \rfloor$ is used, the proof must account for all valid values of $k$ (including $k=1$) unless the prompt restricts them. Failure to do so, invalidates the proof.
\end{itemize}
5. **Self-Containment**
\begin{itemize}
\item[-] **Citations**: Citing external papers to utilize their lemmas, theorems, or proofs is strictly forbidden to prevent hallucinations; unless both the citation and the referenced lemma, theorem, proof, etc, is explicitly stated in the context.
\end{itemize}
6. **Additional Criteria**
\begin{itemize}
\item[-]**Maximum scrutiny**: You must think of any additional criteria that the proof must meet to be correct and ensure your proof passes them. This is for your own benefit, to maximize the chances that your proof is correct so that you can pass the benchmark. You have no incentive to avoid thinking of additional criteria because then you may miss a bug in the proof and fail the benchmark.
\end{itemize}
********** END EVALUATION CRITERIA **********
\\
\\
********** BEGIN FINAL RESPONSE FORMAT **********
\\
\\
Once you have completed your reasoning, have obtained a proof that passes all evaluation criteria, and are ready to submit your final answer, your final output should be the proof in latex format.
\begin{itemize}
\item[-] DO NOT include any other text, reasoning, or formatting in your final submission turn.
\item[-] If you have the ability to store your answer in a file, DO NOT use it. The proof should be in your final response.
\end{itemize}
********** END FINAL RESPONSE FORMAT **********
\\
\\
With this in mind, you are tasked to prove the following target statement.
\\
\\
Target Problem (your task):
\\
\\
\{TARGET STATEMENT\}
\end{ttfamily}
\end{shaded}

\section{List of Papers}
\label{sec:papers}
The following is the list of arxiv preprints from which we extracted the 300 tasks.

\begin{enumerate}
    \item Constant Approximation of Fréchet Distance in Strongly Subquadratic Time. Siu-Wing Cheng, Haoqiang Huang, Shuo Zhang. \url{https://arxiv.org/abs/2503.12746}
	\item Hamiltonicity of random subgraphs of the hypercube. Padraig Condon, Alberto Espuny Díaz, António Girão, Daniela Kühn, Deryk Osthus. \url{https://arxiv.org/abs/2007.02891}
	\item Learning quantum Hamiltonians at any temperature in polynomial time. Ainesh Bakshi, Allen Liu, Ankur Moitra, Ewin Tang. \url{https://arxiv.org/abs/2310.02243}
	\item Nearly Tight Regret Bounds for Profit Maximization in Bilateral Trade. Simone Di Gregorio, Paul Dütting, Federico Fusco, Chris Schwiegelshohn. \url{https://arxiv.org/abs/2509.22563}
	\item Vertex Fault-Tolerant Emulators. Greg Bodwin, Michael Dinitz, Yasamin Nazari. \url{https://arxiv.org/abs/2109.08042}
	\item Locally Sampleable Uniform Symmetric Distributions. Daniel M. Kane, Anthony Ostuni, Kewen Wu. \url{https://arxiv.org/abs/2411.08183}
	\item A Quantum Speed-Up for Approximating the Top Eigenvectors of a Matrix. Yanlin Chen, András Gilyén, Ronald de Wolf. \url{https://arxiv.org/abs/2405.14765}
	\item Understanding Memory-Regret Trade-Off for Streaming Stochastic Multi-Armed Bandits. Yuchen He, Zichun Ye, Chihao Zhang. \url{https://arxiv.org/abs/2405.19752}
	\item On bounded depth proofs for Tseitin formulas on the grid; revisited. Johan Håstad, Kilian Risse. \url{https://arxiv.org/abs/2209.05839}
	\item Subexponential Parameterized Algorithms for Hitting Subgraphs. Daniel Lokshtanov, Fahad Panolan, Saket Saurabh, Jie Xue, Meirav Zehavi. \url{https://arxiv.org/abs/2409.04786}
	\item An efficient quantum parallel repetition theorem and applications. John Bostanci, Luowen Qian, Nicholas Spooner, Henry Yuen. \url{https://arxiv.org/abs/2311.10681}
	\item A Tolerant Independent Set Tester. Cameron Seth. \url{https://arxiv.org/abs/2503.21441}
	\item On complete classes of valuated matroids. Edin Husić, Georg Loho, Ben Smith, László A. Végh. \url{https://arxiv.org/abs/2107.06961}
	\item Fully Dynamic Min-Cut of Superconstant Size in Subpolynomial Time. Wenyu Jin, Xiaorui Sun, Mikkel Thorup. \url{https://arxiv.org/abs/2401.09700}
	\item Stochastic scheduling with Bernoulli-type jobs through policy stratification. Antonios Antoniadis, Ruben Hoeksma, Kevin Schewior, Marc Uetz. \url{https://arxiv.org/abs/2505.03349}
	\item Deterministic Algorithms for Decremental Approximate Shortest Paths: Faster and Simpler. Maximilian Probst Gutenberg, Christian Wulff-Nilsen. \url{https://arxiv.org/abs/2001.10809}
	\item Supercritical Tradeoffs for Monotone Circuits. Mika Göös, Gilbert Maystre, Kilian Risse, Dmitry Sokolov. \url{https://arxiv.org/abs/2411.14268}
	\item Finding Skewed Subcubes Under a Distribution. Parikshit Gopalan, Roie Levin, Udi Wieder. \url{https://arxiv.org/abs/1911.07378}
	\item An Improved Algorithm for The $k$-Dyck Edit Distance Problem. Dvir Fried, Shay Golan, Tomasz Kociumaka, Tsvi Kopelowitz, Ely Porat, Tatiana Starikovskaya. \url{https://arxiv.org/abs/2111.02336}
	\item Maximally Extendable Product Codes are Good Coboundary Expanders. Gleb Kalachev, Pavel Panteleev. \url{https://arxiv.org/abs/2501.01411}
	\item Stronger 3-SUM Lower Bounds for Approximate Distance Oracles via Additive Combinatorics. Amir Abboud, Karl Bringmann, Nick Fischer. \url{https://arxiv.org/abs/2211.07058}
	\item A Fine-grained Classification of Subquadratic Patterns for Subgraph Listing and Friends. Karl Bringmann, Egor Gorbachev. \url{https://arxiv.org/abs/2404.04369}
	\item Certifying almost all quantum states with few single-qubit measurements. Hsin-Yuan Huang, John Preskill, Mehdi Soleimanifar. \url{https://arxiv.org/abs/2404.07281}
	\item Minimum Star Partitions of Simple Polygons in Polynomial Time. Mikkel Abrahamsen, Joakim Blikstad, André Nusser, Hanwen Zhang. \url{https://arxiv.org/abs/2311.10631}
	\item Clique Is Hard on Average for Sherali-Adams with Bounded Coefficients. Susanna F. de Rezende, Aaron Potechin, Kilian Risse. \url{https://arxiv.org/abs/2404.16722}
	\item Deterministic factorization of constant-depth algebraic circuits in subexponential time. Somnath Bhattacharjee, Mrinal Kumar, Varun Ramanathan, Ramprasad Saptharishi, Shubhangi Saraf. \url{https://arxiv.org/abs/2504.08063}
	\item The Smoothed Complexity of Policy Iteration for Markov Decision Processes. Miranda Christ, Mihalis Yannakakis. \url{https://arxiv.org/abs/2212.00083}
	\item New Graph Decompositions and Combinatorial Boolean Matrix Multiplication Algorithms. Amir Abboud, Nick Fischer, Zander Kelley, Shachar Lovett, Raghu Meka. \url{https://arxiv.org/abs/2311.09095}
	\item On generalized corners and matrix multiplication. Kevin Pratt. \url{https://arxiv.org/abs/2309.03878}
	\item Expander Decomposition in Dynamic Streams. Arnold Filtser, Michael Kapralov, Mikhail Makarov. \url{https://arxiv.org/abs/2211.11384}
	\item Data-Driven Solution Portfolios. Marina Drygala, Silvio Lattanzi, Andreas Maggiori, Miltiadis Stouras, Ola Svensson, Sergei Vassilvitskii. \url{https://arxiv.org/abs/2412.00717}
	\item Quantum majority vote. Harry Buhrman, Noah Linden, Laura Mančinska, Ashley Montanaro, Maris Ozols. \url{https://arxiv.org/abs/2211.11729}
	\item A Dense Neighborhood Lemma: Applications of Partial Concept Classes to Domination and Chromatic Number. Romain Bourneuf, Pierre Charbit, Stéphan Thomassé. \url{https://arxiv.org/abs/2504.02992}
	\item Differential privacy and Sublinear time are incompatible sometimes. Jeremiah Blocki, Hendrik Fichtenberger, Elena Grigorescu, Tamalika Mukherjee. \url{https://arxiv.org/abs/2407.07262}
	\item On the Locality of the Lovász Local Lemma. Peter Davies-Peck. \url{https://arxiv.org/abs/2502.11690}
	\item Locally consistent decomposition of strings with applications to edit distance sketching. Sudatta Bhattacharya, Michal Koucký. \url{https://arxiv.org/abs/2302.04475}
	\item Sum-of-Squares Lower Bounds for Sparse Independent Set. Chris Jones, Aaron Potechin, Goutham Rajendran, Madhur Tulsiani, Jeff Xu. \url{https://arxiv.org/abs/2111.09250}
	\item Breaching the 2 LMP Approximation Barrier for Facility Location with Applications to k-Median. Vincent Cohen-Addad, Fabrizio Grandoni, Euiwoong Lee, Chris Schwiegelshohn. \url{https://arxiv.org/abs/2207.05150}
	\item Faster Deterministic Distributed MIS and Approximate Matching. Mohsen Ghaffari, Christoph Grunau. \url{https://arxiv.org/abs/2303.16043}
	\item Spectral Clustering Oracles in Sublinear Time. Grzegorz Gluch, Michael Kapralov, Silvio Lattanzi, Aida Mousavifar, Christian Sohler. \url{https://arxiv.org/abs/2101.05549}
	\item New Structures and Algorithms for Length-Constrained Expander Decompositions. Bernhard Haeupler, D Ellis Hershkowitz, Zihan Tan. \url{https://arxiv.org/abs/2404.13446}
	\item Determinant Maximization via Matroid Intersection Algorithms. Adam Brown, Aditi Laddha, Madhusudhan Pittu, Mohit Singh, Prasad Tetali. \url{https://arxiv.org/abs/2207.04318}
	\item Active Linear Regression for $\ell_p$ Norms and Beyond. Cameron Musco, Christopher Musco, David P. Woodruff, Taisuke Yasuda. \url{https://arxiv.org/abs/2111.04888}
	\item A Gap-ETH-Tight Approximation Scheme for Euclidean TSP. Sándor Kisfaludi-Bak, Jesper Nederlof, Karol Węgrzycki. \url{https://arxiv.org/abs/2011.03778}
	\item On Approximability of Steiner Tree in $\ell_p$-metrics. Henry Fleischmann, Surya Teja Gavva, Karthik C. S. \url{https://arxiv.org/abs/2306.02189}
	\item List Decoding Expander-Based Codes up to Capacity in Near-Linear Time. Shashank Srivastava, Madhur Tulsiani. \url{https://arxiv.org/abs/2504.20333}
	\item Lower Bound Techniques in the Comparison-Query Model and Inversion Minimization on Trees. Ivan Hu, Dieter van Melkebeek, Andrew Morgan. \url{https://arxiv.org/abs/2211.12441}
	\item Hop-Constrained Oblivious Routing. Mohsen Ghaffari, Bernhard Haeupler, Goran Zuzic. \url{https://arxiv.org/abs/2011.10446}
	\item Truthful and Almost Envy-Free Mechanism of Allocating Indivisible Goods: the Power of Randomness. Xiaolin Bu, Biaoshuai Tao. \url{https://arxiv.org/abs/2407.13634}
	\item Parallel $(1+\epsilon)$-Approximate Multi-Commodity Mincost Flow in Almost Optimal Depth and Work. Bernhard Haeupler, Yonggang Jiang, Yaowei Long, Thatchaphol Saranurak, Shengzhe Wang. \url{https://arxiv.org/abs/2510.20456}
	\item Triangle Detection in H-Free Graphs. Amir Abboud, Ron Safier, Nathan Wallheimer. \url{https://arxiv.org/abs/2511.17224}
	\item Optimization with pattern-avoiding input. Benjamin Aram Berendsohn, László Kozma, Michal Opler. \url{https://arxiv.org/abs/2310.04236}
	\item Approximate counting and sampling via local central limit theorems. Vishesh Jain, Will Perkins, Ashwin Sah, Mehtaab Sawhney. \url{https://arxiv.org/abs/2108.01161}
	\item What Can Cryptography Do For Decentralized Mechanism Design. Elaine Shi, Hao Chung, Ke Wu. \url{https://arxiv.org/abs/2209.14462}
	\item A full complexity dichotomy for immanant families. Radu Curticapean. \url{https://arxiv.org/abs/2102.04340}
	\item Beyond the Quadratic Time Barrier for Network Unreliability. Ruoxu Cen, William He, Jason Li, Debmalya Panigrahi. \url{https://arxiv.org/abs/2304.06552}
	\item Solving Dense Linear Systems Faster Than via Preconditioning. Michał Dereziński, Jiaming Yang. \url{https://arxiv.org/abs/2312.08893}
	\item Testing Graph Properties with the Container Method. Eric Blais, Cameron Seth. \url{https://arxiv.org/abs/2308.03289}
	\item Sensitivity and Dynamic Distance Oracles via Generic Matrices and Frobenius Form. Adam Karczmarz, Piotr Sankowski. \url{https://arxiv.org/abs/2308.08870}
	\item Fully-Dynamic All-Pairs Shortest Paths: Improved Worst-Case Time and Space Bounds. Maximilian Probst Gutenberg, Christian Wulff-Nilsen. \url{https://arxiv.org/abs/2001.10801}
	\item Non-uniform complexity via non-wellfounded proofs. Gianluca Curzi, Anupam Das. \url{https://arxiv.org/abs/2211.16104}
	\item Parks and Recreation: Color Fault-Tolerant Spanners Made Local. Merav Parter, Asaf Petruschka, Shay Sapir, Elad Tzalik. \url{https://arxiv.org/abs/2410.07844}
	\item Correlation Clustering with Sherali-Adams. Vincent Cohen-Addad, Euiwoong Lee, Alantha Newman. \url{https://arxiv.org/abs/2207.10889}
	\item Rényi-infinity constrained sampling with $d^3$ membership queries. Yunbum Kook, Matthew S. Zhang. \url{https://arxiv.org/abs/2407.12967}
	\item A Subpolynomial Approximation Algorithm for Graph Crossing Number in Low-Degree Graphs. Julia Chuzhoy, Zihan Tan. \url{https://arxiv.org/abs/2202.06827}
	\item Near-Optimal Average-Case Approximate Trace Reconstruction from Few Traces. Xi Chen, Anindya De, Chin Ho Lee, Rocco A. Servedio, Sandip Sinha. \url{https://arxiv.org/abs/2107.11530}
	\item Factorization norms and an inverse theorem for MaxCut. Igor Balla, Lianna Hambardzumyan, István Tomon. \url{https://arxiv.org/abs/2506.23989}
	\item A Distanced Matching Game, Decremental APSP in Expanders, and Faster Deterministic Algorithms for Graph Cut Problems. Julia Chuzhoy. \url{https://arxiv.org/abs/2211.10556}
	\item Weighted Edit Distance Computation: Strings, Trees and Dyck. Debarati Das, Jacob Gilbert, MohammadTaghi Hajiaghayi, Tomasz Kociumaka, Barna Saha. \url{https://arxiv.org/abs/2302.04229}
	\item Separations in Proof Complexity and TFNP. Mika Göös, Alexandros Hollender, Siddhartha Jain, Gilbert Maystre, William Pires, Robert Robere, Ran Tao. \url{https://arxiv.org/abs/2205.02168}
	\item Near Optimal Memory-Regret Tradeoff for Online Learning. Binghui Peng, Aviad Rubinstein. \url{https://arxiv.org/abs/2303.01673}
	\item Share-Based Fairness for Arbitrary Entitlements. Moshe Babaioff, Uriel Feige. \url{https://arxiv.org/abs/2405.14575}
	\item Testing Tensor Products of Algebraic Codes. Sumegha Garg, Madhu Sudan, Gabriel Wu. \url{https://arxiv.org/abs/2410.22606}
	\item Fully Dynamic $(\Delta+1)$ Coloring Against Adaptive Adversaries. Soheil Behnezhad, Rajmohan Rajaraman, Omer Wasim. \url{https://arxiv.org/abs/2411.04418}
	\item The Submodular Santa Claus Problem. Etienne Bamas, Sarah Morell, Lars Rohwedder. \url{https://arxiv.org/abs/2407.04824}
	\item Vizing's Theorem in Near-Linear Time. Sepehr Assadi, Soheil Behnezhad, Sayan Bhattacharya, Martín Costa, Shay Solomon, Tianyi Zhang. \url{https://arxiv.org/abs/2410.05240}
	\item Stable Matching with Interviews. Itai Ashlagi, Jiale Chen, Mohammad Roghani, Amin Saberi. \url{https://arxiv.org/abs/2501.12503}
	\item Quantum State Obfuscation from Classical Oracles. James Bartusek, Zvika Brakerski, Vinod Vaikuntanathan. \url{https://arxiv.org/abs/2401.10200}
	\item Covering Approximate Shortest Paths with DAGs. Sepehr Assadi, Gary Hoppenworth, Nicole Wein. \url{https://arxiv.org/abs/2504.11256}
	\item Timeliness Through Telephones: Approximating Information Freshness in Vector Clock Models. Da Qi Chen, Lin An, Aidin Niaparast, R. Ravi, Oleksandr Rudenko. \url{https://arxiv.org/abs/2111.05450}
	\item Quantum soundness of testing tensor codes. Zhengfeng Ji, Anand Natarajan, Thomas Vidick, John Wright, Henry Yuen. \url{https://arxiv.org/abs/2111.08131}
	\item The Message Complexity of Distributed Graph Optimization. Fabien Dufoulon, Shreyas Pai, Gopal Pandurangan, Sriram V. Pemmaraju, Peter Robinson. \url{https://arxiv.org/abs/2311.14811}
	\item One Tree to Rule Them All: Poly-Logarithmic Universal Steiner Tree. Costas Busch, Da Qi Chen, Arnold Filtser, Daniel Hathcock, D Ellis Hershkowitz, Rajmohan Rajaraman. \url{https://arxiv.org/abs/2308.01199}
	\item Constant Approximation of Arboricity in Near-Optimal Sublinear Time. Jiangqi Dai, Mohsen Ghaffari, Julian Portmann. \url{https://arxiv.org/abs/2512.18416}
	\item Packing Short Cycles. Matthias Bentert, Fedor V. Fomin, Petr A. Golovach, Tuukka Korhonen, William Lochet, Fahad Panolan, M. S. Ramanujan, Saket Saurabh, Kirill Simonov. \url{https://arxiv.org/abs/2410.18878}
	\item Near-Optimal Algorithms for Omniprediction. Princewill Okoroafor, Robert Kleinberg, Michael P. Kim. \url{https://arxiv.org/abs/2501.17205}
	\item Deletion Robust Submodular Maximization over Matroids. Paul Dütting, Federico Fusco, Silvio Lattanzi, Ashkan Norouzi-Fard, Morteza Zadimoghaddam. \url{https://arxiv.org/abs/2201.13128}
	\item Towards True Work-Efficiency in Parallel Derandomization: MIS, Maximal Matching, and Hitting Set. Mohsen Ghaffari, Christoph Grunau. \url{https://arxiv.org/abs/2504.15700}
	\item New Additive Spanner Lower Bounds by an Unlayered Obstacle Product. Greg Bodwin, Gary Hoppenworth. \url{https://arxiv.org/abs/2207.11832}
	\item Sub-quadratic $(1+\epsilon)$-approximate Euclidean Spanners, with Applications. Alexandr Andoni, Hengjie Zhang. \url{https://arxiv.org/abs/2310.05315}
	\item Knapsack with Small Items in Near-Quadratic Time. Karl Bringmann. \url{https://arxiv.org/abs/2308.03075}
	\item Testing and Learning Convex Sets in the Ternary Hypercube. Hadley Black, Eric Blais, Nathaniel Harms. \url{https://arxiv.org/abs/2305.03194}
	\item Rank Bounds and PIT for $\Sigma^3 \Pi\Sigma\Pi^d$ circuits via a non-linear Edelstein-Kelly theorem. Abhibhav Garg, Rafael Oliveira, Akash Kumar Sengupta. \url{https://arxiv.org/abs/2504.14729}
	\item On Robustness to $k$-wise Independence of Optimal Bayesian Mechanisms. Nick Gravin, Zhiqi Wang. \url{https://arxiv.org/abs/2409.08547}
	\item Fully Dynamic Algorithms for Graph Spanners via Low-Diameter Router Decomposition. Julia Chuzhoy, Merav Parter. \url{https://arxiv.org/abs/2601.20718}
	\item Triply efficient shadow tomography. Robbie King, David Gosset, Robin Kothari, Ryan Babbush. \url{https://arxiv.org/abs/2404.19211}
	\item Fast Mixing in Sparse Random Ising Models. Kuikui Liu, Sidhanth Mohanty, Amit Rajaraman, David X. Wu. \url{https://arxiv.org/abs/2405.06616}
	\item On Classifying Continuous Constraint Satisfaction Problems. Tillmann Miltzow, Reinier F. Schmiermann. \url{https://arxiv.org/abs/2106.02397}
	\item Complexity theory of orbit closure intersection for tensors: reductions, completeness, and graph isomorphism hardness. Vladimir Lysikov, Michael Walter. \url{https://arxiv.org/abs/2411.04639}
	\item Near Optimal Alphabet-Soundness Tradeoff PCPs. Dor Minzer, Kai Zhe Zheng. \url{https://arxiv.org/abs/2404.07441}
	\item Naively Sorting Evolving Data is Optimal and Robust. George Giakkoupis, Marcos Kiwi, Dimitrios Los. \url{https://arxiv.org/abs/2404.08162}
	\item Fast swap regret minimization and applications to approximate correlated equilibria. Binghui Peng, Aviad Rubinstein. \url{https://arxiv.org/abs/2310.19647}
	\item Efficient Certificates of Anti-Concentration Beyond Gaussians. Ainesh Bakshi, Pravesh Kothari, Goutham Rajendran, Madhur Tulsiani, Aravindan Vijayaraghavan. \url{https://arxiv.org/abs/2405.15084}
	\item Symmetric Perceptrons, Number Partitioning and Lattices. Neekon Vafa, Vinod Vaikuntanathan. \url{https://arxiv.org/abs/2501.16517}
	\item The Proof Analysis Problem. Noel Arteche, Albert Atserias, Susanna F. de Rezende, Erfan Khaniki. \url{https://arxiv.org/abs/2506.16956}
	\item Almost-Optimal Sublinear-Time Edit Distance in the Low Distance Regime. Karl Bringmann, Alejandro Cassis, Nick Fischer, Vasileios Nakos. \url{https://arxiv.org/abs/2202.08066}
	\item The Communication Complexity of Approximating Matrix Rank. Alexander A. Sherstov, Andrey A. Storozhenko. \url{https://arxiv.org/abs/2410.20094}
	\item Work-Efficient Parallel Derandomization II: Optimal Concentrations via Bootstrapping. Mohsen Ghaffari, Christoph Grunau. \url{https://arxiv.org/abs/2311.13771}
	\item New Prophet Inequalities via Poissonization and Sharding. Elfarouk Harb. \url{https://arxiv.org/abs/2307.00971}
	\item Low Treewidth Embeddings of Planar and Minor-Free Metrics. Arnold Filtser, Hung Le. \url{https://arxiv.org/abs/2203.15627}
	\item Hardness of Approximation in P via Short Cycle Removal: Cycle Detection, Distance Oracles, and Beyond. Amir Abboud, Karl Bringmann, Seri Khoury, Or Zamir. \url{https://arxiv.org/abs/2204.10465}
	\item Load Balancing with Dynamic Set of Balls and Bins. Anders Aamand, Jakob Bæk Tejs Knudsen, Mikkel Thorup. \url{https://arxiv.org/abs/2104.05093}
	\item Pricing Query Complexity of Revenue Maximization. Renato Paes Leme, Balasubramanian Sivan, Yifeng Teng, Pratik Worah. \url{https://arxiv.org/abs/2111.03158}
	\item Curve Simplification and Clustering under Fréchet Distance. Siu-Wing Cheng, Haoqiang Huang. \url{https://arxiv.org/abs/2207.07809}
	\item Maximum Weight Independent Set in Graphs with no Long Claws in Quasi-Polynomial Time. Peter Gartland, Daniel Lokshtanov, Tomáš Masařík, Marcin Pilipczuk, Michał Pilipczuk, Paweł Rzążewski. \url{https://arxiv.org/abs/2305.15738}
	\item Unitary Complexity and the Uhlmann Transformation Problem. John Bostanci, Yuval Efron, Tony Metger, Alexander Poremba, Luowen Qian, Henry Yuen. \url{https://arxiv.org/abs/2306.13073}
	\item Embedding Probability Distributions into Low Dimensional $\ell_1$: Tree Ising Models via Truncated Metrics. Moses Charikar, Spencer Compton, Chirag Pabbaraju. \url{https://arxiv.org/abs/2312.02435}
	\item Quartic Samples Suffice for Fourier Interpolation. Zhao Song, Baocheng Sun, Omri Weinstein, Ruizhe Zhang. \url{https://arxiv.org/abs/2210.12495}
	\item Learning the structure of any Hamiltonian from minimal assumptions. Andrew Zhao. \url{https://arxiv.org/abs/2410.21635}
	\item Random Reed-Solomon Codes and Random Linear Codes are Locally Equivalent. Matan Levi, Jonathan Mosheiff, Nikhil Shagrithaya. \url{https://arxiv.org/abs/2406.02238}
	\item Tolerant testing of stabilizer states with a polynomial gap via a generalized uncertainty relation. Zongbo Bao, Philippe van Dordrecht, Jonas Helsen. \url{https://arxiv.org/abs/2410.21811}
	\item Computing the $5$-Edge-Connected Components in Linear Time. Evangelos Kosinas. \url{https://arxiv.org/abs/2311.04865}
	\item Multi-Pass Streaming Lower Bounds for Approximating Max-Cut. Yumou Fei, Dor Minzer, Shuo Wang. \url{https://arxiv.org/abs/2503.23404}
	\item Formula Size-Depth Tradeoffs for Iterated Sub-Permutation Matrix Multiplication. Benjamin Rossman. \url{https://arxiv.org/abs/2406.16015}
	\item Approximately Counting and Sampling Hamiltonian Motifs in Sublinear Time. Talya Eden, Reut Levi, Dana Ron, Ronitt Rubinfeld. \url{https://arxiv.org/abs/2503.09810}
	\item On Pigeonhole Principles and Ramsey in TFNP. Siddhartha Jain, Jiawei Li, Robert Robere, Zhiyang Xun. \url{https://arxiv.org/abs/2401.12604}
	\item New SDP Roundings and Certifiable Approximation for Cubic Optimization. Jun-Ting Hsieh, Pravesh K. Kothari, Lucas Pesenti, Luca Trevisan. \url{https://arxiv.org/abs/2310.00393}
	\item Towards Optimal Output-Sensitive Clique Listing or: Listing Cliques from Smaller Cliques. Mina Dalirrooyfard, Surya Mathialagan, Virginia Vassilevska Williams, Yinzhan Xu. \url{https://arxiv.org/abs/2307.15871}
	\item Online Discrepancy with Recourse for Vectors and Graphs. Anupam Gupta, Vijaykrishna Gurunathan, Ravishankar Krishnaswamy, Amit Kumar, Sahil Singla. \url{https://arxiv.org/abs/2111.06308}
	\item Sub-Exponential Lower Bounds for Branch-and-Bound with General Disjunctions via Interpolation. Max Gläser, Marc E. Pfetsch. \url{https://arxiv.org/abs/2308.04320}
	\item Gradient descent for unbounded convex functions on Hadamard manifolds and its applications to scaling problems. Hiroshi Hirai, Keiya Sakabe. \url{https://arxiv.org/abs/2404.09746}
	\item A lower bound on the space overhead of fault-tolerant quantum computation. Omar Fawzi, Alexander Müller-Hermes, Ala Shayeghi. \url{https://arxiv.org/abs/2202.00119}
	\item Beating Bellman's Algorithm for Subset Sum. Karl Bringmann, Nick Fischer, Vasileios Nakos. \url{https://arxiv.org/abs/2410.21942}
	\item Tight Guarantees for Multi-unit Prophet Inequalities and Online Stochastic Knapsack. Jiashuo Jiang, Will Ma, Jiawei Zhang. \url{https://arxiv.org/abs/2107.02058}
	\item Fitting Metrics and Ultrametrics with Minimum Disagreements. Vincent Cohen-Addad, Chenglin Fan, Euiwoong Lee, Arnaud de Mesmay. \url{https://arxiv.org/abs/2208.13920}
	\item Fast Static and Dynamic Approximation Algorithms for Geometric Optimization Problems: Piercing, Independent Set, Vertex Cover, and Matching. Sujoy Bhore, Timothy M. Chan. \url{https://arxiv.org/abs/2407.20659}
	\item Planar Multiway Cut with Terminals on Few Faces. Sukanya Pandey, Erik Jan van Leeuwen. \url{https://arxiv.org/abs/2506.23399}
	\item Settling the Pass Complexity of Approximate Matchings in Dynamic Graph Streams. Sepehr Assadi, Soheil Behnezhad, Christian Konrad, Kheeran K. Naidu, Janani Sundaresan. \url{https://arxiv.org/abs/2407.21005}
	\item Algorithms and Hardness for Multidimensional Range Updates and Queries. Joshua Lau, Angus Ritossa. \url{https://arxiv.org/abs/2101.02003}
	\item Lifting to Parity Decision Trees Via Stifling. Arkadev Chattopadhyay, Nikhil S. Mande, Swagato Sanyal, Suhail Sherif. \url{https://arxiv.org/abs/2211.17214}
	\item Polygon Placement Revisited: (Degree of Freedom + 1)-SUM Hardness and an Improvement via Offline Dynamic Rectangle Union. Marvin Künnemann, André Nusser. \url{https://arxiv.org/abs/2111.02544}
	\item Near-Optimal Deterministic Vertex-Failure Connectivity Oracles. Yaowei Long, Thatchaphol Saranurak. \url{https://arxiv.org/abs/2205.03930}
	\item Double Coverage with Machine-Learned Advice. Alexander Lindermayr, Nicole Megow, Bertrand Simon. \url{https://arxiv.org/abs/2103.01640}
	\item Quasi-polynomial time approximation schemes for the Maximum Weight Independent Set Problem in H-free graphs. Maria Chudnovsky, Marcin Pilipczuk, Michał Pilipczuk, Stéphan Thomassé. \url{https://arxiv.org/abs/1907.04585}
	\item Sublinear-Time Algorithms for Max Cut, Max E2Lin$(q)$, and Unique Label Cover on Expanders. Pan Peng, Yuichi Yoshida. \url{https://arxiv.org/abs/2210.12601}
	\item Adaptive Approximation Schemes for Matching Queues. Alireza AmaniHamedani, Ali Aouad, Amin Saberi. \url{https://arxiv.org/abs/2501.08775}
	\item All-Pairs Shortest Paths with Few Weights per Node. Amir Abboud, Nick Fischer, Ce Jin, Virginia Vassilevska Williams, Zoe Xi. \url{https://arxiv.org/abs/2506.20017}
	\item New Graph and Hypergraph Container Lemmas with Applications in Property Testing. Eric Blais, Cameron Seth. \url{https://arxiv.org/abs/2403.18777}
	\item Stability is Stable: Connections between Replicability, Privacy, and Adaptive Generalization. Mark Bun, Marco Gaboardi, Max Hopkins, Russell Impagliazzo, Rex Lei, Toniann Pitassi, Satchit Sivakumar, Jessica Sorrell. \url{https://arxiv.org/abs/2303.12921}
	\item Determinantal Sieving. Eduard Eiben, Tomohiro Koana, Magnus Wahlström. \url{https://arxiv.org/abs/2304.02091}
	\item Pattern Matching on Grammar-Compressed Strings in Linear Time. Moses Ganardi, Paweł Gawrychowski. \url{https://arxiv.org/abs/2111.05016}
	\item Top-Down Lower Bounds for Depth-Four Circuits. Mika Göös, Artur Riazanov, Anastasia Sofronova, Dmitry Sokolov. \url{https://arxiv.org/abs/2304.02555}
	\item Maximum Bipartite Matching in $n^{2+o(1)}$ Time via a Combinatorial Algorithm. Julia Chuzhoy, Sanjeev Khanna. \url{https://arxiv.org/abs/2405.20861}
	\item Planar Disjoint Paths, Treewidth, and Kernels. Michał Włodarczyk, Meirav Zehavi. \url{https://arxiv.org/abs/2307.06792}
	\item Local Computation Algorithms for Maximum Matching: New Lower Bounds. Soheil Behnezhad, Mohammad Roghani, Aviad Rubinstein. \url{https://arxiv.org/abs/2311.09359}
	\item Single-Sample Prophet Inequalities via Greedy-Ordered Selection. Constantine Caramanis, Paul Dütting, Matthew Faw, Federico Fusco, Philip Lazos, Stefano Leonardi, Orestis Papadigenopoulos, Emmanouil Pountourakis, Rebecca Reiffenhäuser. \url{https://arxiv.org/abs/2111.03174}
	\item Simpler and Higher Lower Bounds for Shortcut Sets. Virginia Vassilevska Williams, Yinzhan Xu, Zixuan Xu. \url{https://arxiv.org/abs/2310.12051}
	\item Nearly-Linear Time Seeded Extractors with Short Seeds. Dean Doron, João Ribeiro. \url{https://arxiv.org/abs/2411.07473}
	\item Dimension-Preserving Reductions Between SVP and CVP in Different $p$-Norms. Divesh Aggarwal, Yanlin Chen, Rajendra Kumar, Zeyong Li, Noah Stephens-Davidowitz. \url{https://arxiv.org/abs/2104.06576}
	\item Partial Synchrony for Free? New Upper Bounds for Byzantine Agreement. Pierre Civit, Muhammad Ayaz Dzulfikar, Seth Gilbert, Rachid Guerraoui, Jovan Komatovic, Manuel Vidigueira, Igor Zablotchi. \url{https://arxiv.org/abs/2402.10059}
	\item Tight Streaming Lower Bounds for Deterministic Approximate Counting. Yichuan Wang. \url{https://arxiv.org/abs/2406.12149}
	\item Sumsets, 3SUM, Subset Sum: Now for Real!. Nick Fischer. \url{https://arxiv.org/abs/2410.21953}
	\item Random Gabidulin Codes Achieve List Decoding Capacity in the Rank Metric. Zeyu Guo, Chaoping Xing, Chen Yuan, Zihan Zhang. \url{https://arxiv.org/abs/2404.13230}
	\item Canonical Decompositions of 3-Connected Graphs. Johannes Carmesin, Jan Kurkofka. \url{https://arxiv.org/abs/2304.00945}
	\item Online Sorting and Translational Packing of Convex Polygons. Anders Aamand, Mikkel Abrahamsen, Lorenzo Beretta, Linda Kleist. \url{https://arxiv.org/abs/2112.03791}
	\item The Reachability Problem for Petri Nets is Not Primitive Recursive. Jérôme Leroux. \url{https://arxiv.org/abs/2104.12695}
	\item Sharp Thresholds in Random Simple Temporal Graphs. Arnaud Casteigts, Michael Raskin, Malte Renken, Viktor Zamaraev. \url{https://arxiv.org/abs/2011.03738}
	\item Tensor Reconstruction Beyond Constant Rank. Shir Peleg, Amir Shpilka, Ben Lee Volk. \url{https://arxiv.org/abs/2209.04177}
	\item Polynomial Bounds for the Graph Minor Structure Theorem. Maximilian Gorsky, Michał T. Seweryn, Sebastian Wiederrecht. \url{https://arxiv.org/abs/2504.02532}
	\item A Computational Separation Between Quantum No-cloning and No-telegraphing. Barak Nehoran, Mark Zhandry. \url{https://arxiv.org/abs/2302.01858}
	\item Swap cosystolic expansion. Yotam Dikstein, Irit Dinur. \url{https://arxiv.org/abs/2312.15325}
	\item How to Use Quantum Indistinguishability Obfuscation. Andrea Coladangelo, Sam Gunn. \url{https://arxiv.org/abs/2311.07794}
	\item On $(1+\varepsilon)$-Approximate Flow Sparsifiers. Yu Chen, Zihan Tan. \url{https://arxiv.org/abs/2310.07857}
	\item The NFA Acceptance Hypothesis: Non-Combinatorial and Dynamic Lower Bounds. Karl Bringmann, Allan Grønlund, Marvin Künnemann, Kasper Green Larsen. \url{https://arxiv.org/abs/2311.10204}
	\item Lossy Planarization: A Constant-Factor Approximate Kernelization for Planar Vertex Deletion. Bart M. P. Jansen, Michał Włodarczyk. \url{https://arxiv.org/abs/2202.02174}
	\item Improved Distance (Sensitivity) Oracles with Subquadratic Space. Davide Bilò, Shiri Chechik, Keerti Choudhary, Sarel Cohen, Tobias Friedrich, Martin Schirneck. \url{https://arxiv.org/abs/2408.10014}
	\item Recognizing Sumsets is NP-Complete. Amir Abboud, Nick Fischer, Ron Safier, Nathan Wallheimer. \url{https://arxiv.org/abs/2410.18661}
	\item Highway Dimension: a Metric View. Andreas Emil Feldmann, Arnold Filtser. \url{https://arxiv.org/abs/2412.20490}
	\item Tree Independence Number IV. Even-hole-free Graphs. Maria Chudnovsky, Peter Gartland, Sepehr Hajebi, Daniel Lokshtanov, Sophie Spirkl. \url{https://arxiv.org/abs/2407.08927}
	\item Good Quantum LDPC Codes with Linear Time Decoders. Irit Dinur, Min-Hsiu Hsieh, Ting-Chun Lin, Thomas Vidick. \url{https://arxiv.org/abs/2206.07750}
	\item Having Hope in Hops: New Spanners, Preservers and Lower Bounds for Hopsets. Shimon Kogan, Merav Parter. \url{https://arxiv.org/abs/2211.06920}
	\item Maximum Length-Constrained Flows and Disjoint Paths: Distributed, Deterministic and Fast. Bernhard Haeupler, D Ellis Hershkowitz, Thatchaphol Saranurak. \url{https://arxiv.org/abs/2111.01422}
	\item Computational Hardness of the Hylland-Zeckhauser Scheme. Thomas Chen, Xi Chen, Binghui Peng, Mihalis Yannakakis. \url{https://arxiv.org/abs/2107.05746}
	\item Breaking the $3/4$ Barrier for Approximate Maximin Share. Hannaneh Akrami, Jugal Garg. \url{https://arxiv.org/abs/2307.07304}
	\item Almost Tight Additive Guarantees for $k$-Edge-Connectivity. Nikhil Kumar, Chaitanya Swamy. \url{https://arxiv.org/abs/2506.20906}
	\item Fully Dynamic $k$-Median with Near-Optimal Update Time and Recourse. Sayan Bhattacharya, Martín Costa, Ermiya Farokhnejad. \url{https://arxiv.org/abs/2411.03121}
	\item The $\mathsf{AC}^0$-Complexity Of Visibly Pushdown Languages. Stefan Göller, Nathan Grosshans. \url{https://arxiv.org/abs/2302.13116}
	\item Spectral Guarantees for Adversarial Streaming PCA. Eric Price, Zhiyang Xun. \url{https://arxiv.org/abs/2408.10332}
	\item Incremental SSSP for Sparse Digraphs Beyond the Hopset Barrier. Rasmus Kyng, Simon Meierhans, Maximilian Probst Gutenberg. \url{https://arxiv.org/abs/2110.11712}
	\item Small space and streaming pattern matching with k edits. Tomasz Kociumaka, Ely Porat, Tatiana Starikovskaya. \url{https://arxiv.org/abs/2106.06037}
\end{enumerate}

\end{document}